\documentclass[10pt,conference]{IEEEtran}
\usepackage[ruled,vlined]{algorithm2e}

\usepackage[nolist]{acronym}
\usepackage{amsmath}

\usepackage{mathtools}
\usepackage{amsthm}
\usepackage[pdftex]{graphicx}

\usepackage[urlcolor=blue]{hyperref} 
\usepackage[table,xcdraw]{xcolor}
\usepackage{graphics}
\usepackage{multirow}
\usepackage{graphicx}
\usepackage[colorinlistoftodos]{todonotes}
\usepackage{pifont}
\usepackage{ dsfont }
\usepackage{amssymb}
\usepackage{multicol}
\usepackage{placeins}

\usepackage{amsmath}
\usepackage[ruled,vlined]{algorithm2e}

\usepackage{needspace}

\usepackage{cite}
\usepackage{amsmath,amssymb,amsfonts}
\usepackage{algorithmic}
\usepackage{graphicx,color}
\usepackage{textcomp}

\usepackage{cite} 
\usepackage[nolist]{acronym}
\usepackage{amsmath}
 \usepackage{booktabs}
\usepackage{mathtools}
\usepackage{amsthm}
\usepackage[pdftex]{graphicx}

\usepackage[urlcolor=blue]{hyperref} 
\usepackage[table,xcdraw]{xcolor}
\usepackage{graphics}
\usepackage{multirow}
\usepackage{multirow}
\usepackage{graphicx}
\usepackage[table,xcdraw]{xcolor}
\usepackage[normalem]{ulem}
\useunder{\uline}{\ul}{}
\usepackage{graphicx}
\usepackage[table,xcdraw]{xcolor}
\usepackage[colorinlistoftodos]{todonotes}
\usepackage{pifont}
\usepackage{svg}
\usepackage{dsfont}
\usepackage{amssymb}
\usepackage{multicol}
\usepackage{placeins}
\usepackage{comment}
\usepackage{tikz}
\usetikzlibrary{shapes,shadows,arrows}
\tikzstyle{decision} = [diamond, draw, fill= blue!50]

\tikzstyle{elli} = [draw, ellipse, fill=red!50, minimum height = 8mm]
\tikzstyle{block} = [draw, rectangle, fill= blue!50, text width=8em, text centered, minimum height = 15mm, node distance=5em]
\tikzstyle{line} = [draw,-latex']

\usepackage{cleveref}
\crefname{figure}{Fig}{Figs}
\Crefname{figure}{Fig}{Figs}
\crefname{table}{Table}{Tables}
\Crefname{table}{Table}{Tables}

\usepackage{tikz}
\usetikzlibrary{shapes,shadows,arrows}
\tikzstyle{decision} = [diamond, draw, fill= blue!50]
\tikzstyle{line} = [draw, -latex']
\tikzstyle{elli} = [draw, ellipse, fill=red!50, minimum height = 8mm]
\tikzstyle{block} = [draw, rectangle, fill= blue!50, text width=8em, text centered, minimum height = 15mm, node distance=5em]
\tikzstyle{line} = [draw,-latex']

\usepackage{comment}

\title{An Explainable AI Framework for Dynamic Resource Management in Vehicular Network Slicing}

\author{
    Haochen Sun\IEEEauthorrefmark{1},~\IEEEmembership{Student Member,~IEEE}
    Yifan Liu\IEEEauthorrefmark{1},~\IEEEmembership{Member,~IEEE} 
    Ahmed Al-Tahmeesschi\IEEEauthorrefmark{1},~\IEEEmembership{Member,~IEEE} Swarna Chetty\IEEEauthorrefmark{1},~\IEEEmembership{Member,~IEEE} Syed Ali Raza Zaidi\IEEEauthorrefmark{2}\\
    Avishek Nag\IEEEauthorrefmark{3},~\IEEEmembership{Senior Member,~IEEE}
    and Hamed Ahmadi\IEEEauthorrefmark{1},~\IEEEmembership{Senior Member,~IEEE}
    
    \\

    \IEEEauthorrefmark{1}Institute for Safe Autonomy and School of Physics, Engineering and
Technology, York, YO10 5DD, UK\\
    \IEEEauthorrefmark{2}School of Electrical and Electronic Engineering, University of Leeds, Leeds, UK\\
    \IEEEauthorrefmark{3}School of Computer Science, University College Dublin, Dublin, Ireland
}

\begin{document}
\maketitle

\begin{abstract}
Effective resource management and network slicing are essential to meet the diverse service demands of vehicular networks, including \ac{eMBB} and \ac{URLLC}. This paper introduces an Explainable Deep Reinforcement Learning (XRL) framework for dynamic network slicing and resource allocation in vehicular networks, built upon a near-real-time RAN intelligent controller. By integrating a feature-based approach that leverages Shapley values and an attention mechanism, we interpret and refine the decisions of our reinforcement learning agents, addressing key reliability challenges in vehicular communication systems. 
Simulation results demonstrate that our approach provides clear, real-time insights into the resource allocation process and achieves higher interpretability precision than a pure attention mechanism. Furthermore, the \ac{QoS} satisfaction for \ac{URLLC} services increased from 78.0\% to 80.13\%, while that for \ac{eMBB} services improved from 71.44\% to 73.21\%.
\end{abstract}


\IEEEpeerreviewmaketitle

\section{Introduction}

Vehicular communication networks are poised to play a central role in \ac{6G} wireless systems, where ultra-high data rates, extremely low latency, and ubiquitous intelligence are envisioned
as key enablers for future mobility and automation. However, achieving reliable real-time communication for highly dynamic vehicular environments remains a major challenge. The increasing complexity of \ac{V2X} services demands efficient and dynamic resource management~\cite{10854503}. Conventional networks often fall short in addressing the challenges of vehicular services, which are marked by high mobility, fluctuating channel conditions, and diverse \ac{QoS} requirements. For instance, safety-critical applications demand ultra-reliable low-latency communications (\ac{URLLC}), while multimedia services rely on enhanced mobile broadband (\ac{eMBB}). In response, network slicing has emerged as a pivotal technology that partitions a single physical infrastructure into multiple, logically isolated networks, each tailored to specific service demands~\cite{khan2020network}.However, the operation of network slicing relies on efficient and precise resource allocation. While traditional mathematical optimization methods prove effective in static and predictable scenarios, they struggle to cope with the inherent dynamics, strict delay constraints, and frequent topology changes typical of vehicular environments~\cite{su2019resource}.

Recent advancements in \ac{DRL} have demonstrated significant promise for dynamically allocation resources in highly variable environments. For instance, algorithms based on \ac{DDPG} have shown both efficiency and adaptability in addressing multi-dimensional resource requirements through coordinated strategies among distributed agents~\cite{cui2023multi}. Similarly, ~\cite{8796358} proposed a multi-agent DRL framework leveraging dueling double deep Q-networks (D3QN) to effectively address joint user association and resource allocation problems in heterogeneous cellular networks, further demonstrating \ac{DRL}'s ability to handle non-convex, combinatorial resource management scenarios. However, despite their high performance, such \ac{DRL}-based approaches often function as closed boxes, lacking the interpretability required to build trust and ensure reliability in safety-critical vehicular applications~\cite{tang2021comprehensive}.

To mitigate these challenges, Explainable AI (\ac{XAI}) techniques have been developed to provide post-hoc explanations that enhance the transparency of model decisions\cite{dwivedi2023explainable}. Among these, the Shapley Values for Explaining Reinforcement Learning (SVERL) method offers a rigorous, game-theoretic foundation to elucidate the contributions of individual features to an agent's performance~\cite{beechey2023explaining}. Nevertheless, the traditional SVERL approach suffers from significant computational overhead, making it unsuitable for real-time applications, especially in latency-sensitive vehicular networks.

To overcome this limitation, we propose an integrated framework that combines an attention mechanism with the SVERL methodology, yielding an efficient and interpretable \ac{DRL} solution tailored for vehicular network slicing. Our approach guarantees high-quality real-time explainability by supervising the attention layer with approximate Shapley values during training. This integration ensures that the trained models can quickly generate both accurate resource allocation decisions and comprehensible explanations.

The primary contributions of this work include the development of a novel interpretable \ac{DRL} framework that fuses an attention-based \ac{DDPG} model with SVERL for dynamic vehicular network slicing and resource management. The proposed solution not only provides an interpretable decision-making process—thereby enhancing trust and transparency in highly dynamic vehicular environments—but also improves \ac{QoS} satisfaction for various \ac{V2X} services.

\section{System Model}
\subsection{Network Topology}
\begin{figure}[h]
    \centering
    \includegraphics[clip, trim=8cm 1cm 8cm 1cm, width=0.35\textwidth]{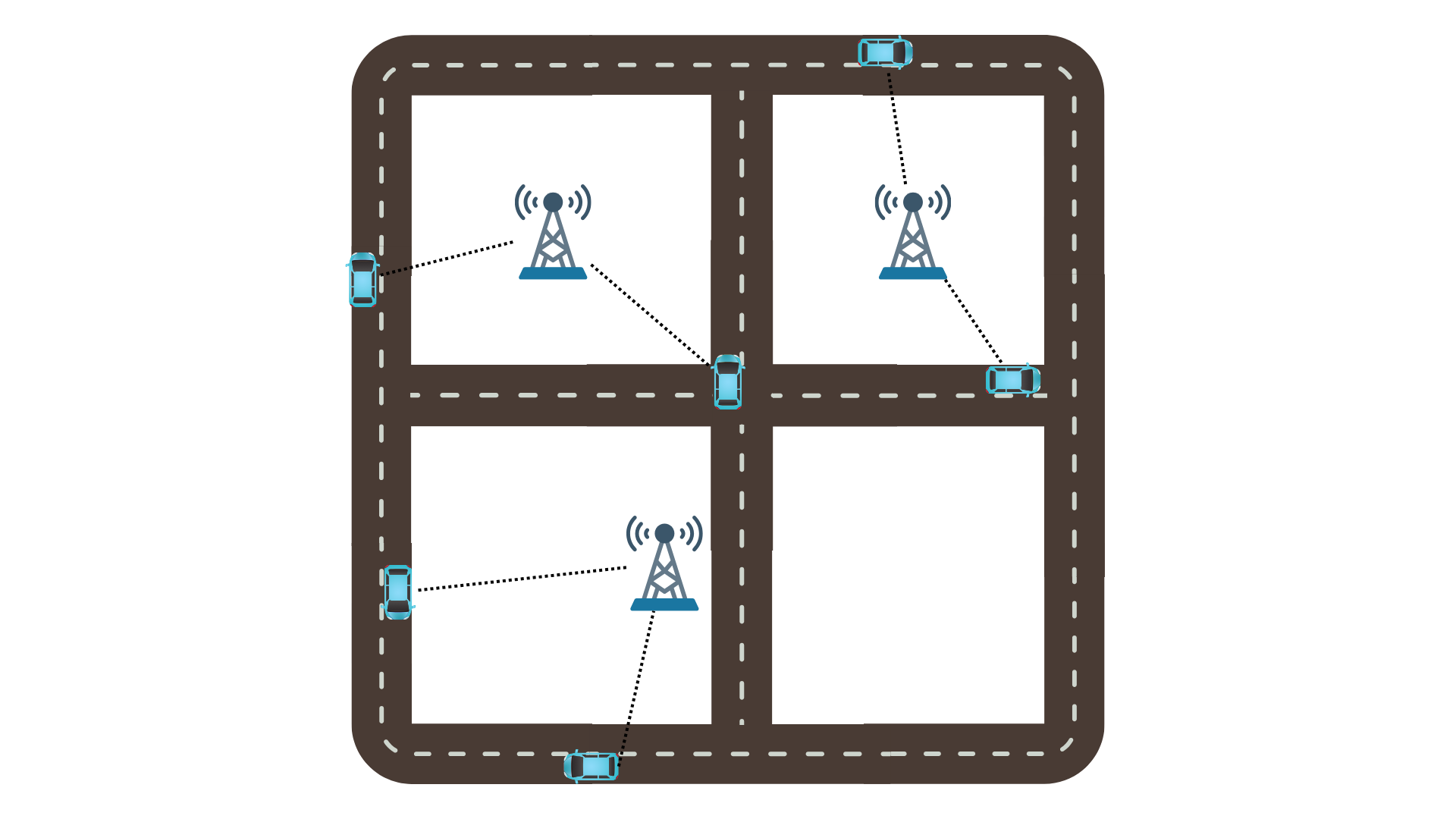}
    \caption{Vehicular Network Environment}
    \label{fig:examplecounter}
\end{figure}

We consider a vehicular network deployed over a grid road topology with four primary intersections, as illustrated in Fig.~\ref{fig:examplecounter}. A set of \acp{gNB}, denoted by $\mathcal{M}_{\text{all}} = {1,2,\dots,M}$, is strategically deployed based on classical deployment strategies aimed at coverage optimization and interference minimization to ensure comprehensive coverage. A set of vehicles, $\mathcal{N} = \{1,2,\dots, N\}$, traverses the network along random paths while requesting services (e.g., \ac{URLLC} or \ac{eMBB}) according to their specific \ac{QoS} demands~\cite{10854503,khan2020network}. Time is discretized into slots $t \in \{1,2,\dots\}$, and the connection between vehicle $i$ and \ac{gNB} $m$ at time $t$ is indicated by the binary variable $q_{i,m}^t \in \{0,1\}$, which satisfies
\begin{equation}
\sum_{m=1}^{M} q_{i,m}^t = 1, \quad \forall i \in \mathcal{N}.
\end{equation}
 This ensures that each vehicle is connected to exactly one \ac{gNB} per time slot. Moreover, we assume that all \acp{gNB} are interconnected via low-latency wired backhaul links to a centralized \ac{SDN} controller, enabling efficient global resource management~\cite{cui2023multi,das2018radio}.

\subsection{Communication Model}
Communication between vehicles and \acp{gNB} is enabled by the allocation of \acp{PRB} rather than direct bandwidth assignment. In 5G NR, a \ac{PRB} comprises 12 consecutive subcarriers, with its bandwidth determined by the subcarrier spacing (e.g., 360 kHz for a 30 kHz spacing)~\cite{3GPP_TS_38.211_V16.3.0}. The instantaneous throughput for vehicle $i$ connected to \ac{gNB} $m$ at time $t$ is given by:
\begin{equation}
R_{i,m}^t = B_{i,m}\cdot 3.6 \times 10^5 \cdot \log_2\left(1 + \frac{P\,G_{i,m}^t}{\sigma^2 + I_m^t}\right),
\label{eq:transmission_rate_modified}
\end{equation}
where $B_{i,m}$ denotes the number of allocated \acp{PRB}, $P$ is the transmission power, $\sigma^2$ represents the noise power, and $I_m^t$ is the interference power. For a vehicle with data demand $D_i^t$, the total communication delay is calculated as:
\begin{equation}
T_{i,m}^t = \frac{D_i^t}{R_{i,m}^t} + T_{\text{processing}} + T_{\text{scheduling}},
\label{eq:communication_delay}
\end{equation}
with $T_{\text{processing}}+T_{\text{scheduling}}$ fixed at 1.5~ms, in accordance with relevant technical standards~\cite{3GPP_TS_22.186_V16.2.0}.

\subsection{Slice Assignment Model}
To address the heterogeneous service requirements in vehicular networks, vehicle services are categorized into two network slices: \ac{URLLC} for latency-sensitive, safety-critical applications and \ac{eMBB} for high data-rate applications~\cite{ITU-R_M.2083-0,khan2021network}. The service requests for each vehicle are denoted by the binary variables $s_{i}^{\text{URLLC}}$ and $s_{i}^{\text{eMBB}}$ which are generated randomly upon the vehicle's entry into the vehicular network environment and remain constant throughout its journey. For vehicles requesting \ac{URLLC} services, the delay must remain below the maximum threshold $T_{\text{th}}$:
\begin{equation}
s_{i}^{\text{URLLC}} \cdot T_{i,m}^t \leq s_{i}^{\text{URLLC}} \cdot T_{\text{th}}, \quad \forall i \in \mathcal{N},
\label{eq:urllc_slice_binary}
\end{equation}
while for vehicles requiring \ac{eMBB} services, the throughput must meet a minimum threshold $R_{\text{th}}$:
\begin{equation}
s_{i}^{\text{eMBB}} \cdot R_{i,m}^t \geq s_{i}^{\text{eMBB}} \cdot R_{\text{th}}, \quad \forall i \in \mathcal{N}.
\label{eq:embb_slice_binary}
\end{equation}
Since vehicles may simultaneously request both services, this flexible slice assignment effectively accommodates diverse \ac{QoS} requirements.

\section{Problem Formulation}
\label{sec:pro}
The primary objective of this study is to minimize the overall \ac{QoS} violations experienced by all vehicles while ensuring that diverse service requirements are met and that the model's resource allocation decisions remain interpretable. In this work, \ac{QoS} violations are quantified as penalties incurred under two conditions: for vehicles in the \ac{URLLC} slice, a penalty is imposed when the communication delay $T_{i,m}^t$ exceeds the maximum allowable latency $T_{\max}$; for vehicles in the \ac{eMBB} slice, a penalty is incurred when the achieved throughput $R_{i,m}^t$ falls below the required threshold $R_{\min}$. Consequently, the optimization problem is formulated as:

\begin{equation}
\begin{aligned}
\min_{\{ B_{i,m}^t,\,q_{i,m}^t \}}
\quad & \sum_{t=1}^{T}\sum_{i=1}^{N}
\Biggl[
s_{i}^{\text{URLLC}}\,
\frac{\max\!\bigl(0,\,T_{i,m}^t - T_{\text{th}}\bigr)}{T_{\text{th}}}
\\
&\qquad\qquad+\;
s_{i}^{\text{eMBB}}\,
\frac{\max\!\bigl(0,\,R_{\text{th}} - R_{i,m}^t\bigr)}{R_{\text{th}}}
\Biggr],
\end{aligned}
\label{eq:obj_function}
\end{equation}
s.t.:
\addtocounter{equation}{-1}
\begin{subequations}\label{eq:obj_functioncons}
\begin{align}
\sum_{i \in \mathcal{N}} B_{i,m}^t &\le W_m,
\quad \forall m,\,t, \label{eq:obj_function1}\\
\sum_{m \in \mathcal{M}} q_{i,m}^t &= 1,\; q_{i,m}^t \in \{0,1\},
\quad \forall i,\,t, \label{eq:obj_function2}\\
B_{i,m}^t &\le W_m\,q_{i,m}^t,
\quad \forall i,\,m,\,t. \label{eq:obj_function3}
\end{align}
\end{subequations}

Constraint~\eqref{eq:obj_function1} ensures that the total number of \acp{PRB} allocated by any \ac{gNB} does not exceed its capacity $W_m$. Constraint~\eqref{eq:obj_function2} enforces that each vehicle is connected to exactly one \ac{gNB} each slot. Finally, Constraint~\eqref{eq:obj_function3} links the resource allocation directly to the active connection. Due to the inherent complexity of vehicular network slicing—marked by dynamic mobility, fluctuating channel conditions, and unpredictable traffic—traditional optimization techniques are often impractical. Thus, we adopt a \ac{DRL} framework that continuously interacts with the network environment, enabling real-time, data-driven resource management without requiring explicit forecasting~\cite{hurtado2022deep}. While our \ac{DRL}-based framework inherently avoids explicit long-term predictions of future network conditions or complicated state forecasting, we utilize readily available short-term mobility information, such as the next timestep vehicle positions, to enhance state representation without incurring significant computational overhead.

\section{MDP Formulation}
\label{sec：MDP}
To address vehicular networks' dynamic and uncertain nature, we formulate the resource allocation and slice assignment problem as a Markov Decision Process (MDP). The key components of the MDP are described below.

\subsubsection{State (\( S \))}
The system state at each discrete time slot $t$ is represented by a comprehensive feature vector that includes both vehicle-specific and network-wide information:
\begin{equation}
\small
\begin{aligned}
S_t = \Bigl\{(x_i^t,\,y_i^t,\,D_i^t,\,s_{i}^{\text{URLLC}},\,s_{i}^{\text{eMBB}},\,a_i^t,\,q_{i,m}^{t-1},\,B_{i,m}^t,\,G_{i,m}^t,\\[-2pt]
x_i^{\text{next},t},\,y_i^{\text{next},t},\,v_i^t) \,\big|\; i \in \mathcal{N},\, m \in \mathcal{M}\Bigr\} \cup \Bigl\{ L_m^t \,\big|\; m \in \mathcal{M} \Bigr\},
\end{aligned}
\normalsize
\label{eq:state_definition}
\end{equation}

where $x_i^t$ and $y_i^t$ are the normalized coordinates of vehicle $i$, $D_i^t$ represents its normalized URLLC data demand, and $s_{i}^{\text{URLLC}}$ and $s_{i}^{\text{eMBB}}$ indicate the service requests. The binary variable $a_i^t$ denotes the activity status of vehicle $i$. Additionally, $q_{i,m}^{t-1}$ and $B_{i,m}^t$ provide historical context by representing the previous vehicle-to-\ac{gNB} association and current resource allocation, respectively. The channel gain $G_{i,m}^t$ reflects the wireless link quality, while $v_i^t$ is speed of vehicle which can capture next time step vehicle's position $x_i^{\text{next},t}$, $y_i^{\text{next},t}$, Finally, $L_m^t$ denotes the normalized load of \ac{gNB} $m$, which assists the agent in anticipating congestion.

\subsubsection{Action (\( A \))}
At each time slot $t$, the RL agent selects an action that comprises two components: the vehicle-to-\ac{gNB} association and the allocation of resource blocks. The action space is defined as:
\begin{equation}
\small
A_t = \Bigl\{(q_{i,m}^t,\,B_{i,m}^t) \,\big|\; i \in \mathcal{N},\, m \in \mathcal{M}\Bigr\},
\label{eq:action_definition}
\normalsize
\end{equation}
where $q_{i,m}^t \in [0,1]$ is a relaxed association variable and $B_{i,m}^t \in [0,W_m]$ represents the allocated resources. While practical systems require binary connectivity and discrete resource allocation, the relaxation facilitates gradient-based optimization during training~\cite{maddison2016concrete}.

\subsubsection{Reward (\( R \))}
The reward function guides the RL agent to minimize \ac{QoS} violations while efficiently utilizing network resources. At each time slot $t$, the immediate reward is defined as:
\begin{equation}
\begin{aligned}
r_t = -\Biggl(
\sum_{i \in \mathcal{N}} s_{i}^{\text{URLLC}}\,\omega_{\text{URLLC}}\,
\frac{\max(0,\,T_{i,m}^t - T_{\text{th}})}{T_{\text{th}}}
\\
+\sum_{i \in \mathcal{N}} s_{i}^{\text{eMBB}}\,\omega_{\text{eMBB}}\,
\frac{\max(0,\,R_{\text{th}} - R_{i,m}^t)}{R_{\text{th}}}
\Biggr),
\end{aligned}
\label{eq:reward_function}
\end{equation}
where $\omega_{\text{URLLC}}$ and $\omega_{\text{eMBB}}$ are weighting factors that emphasize the relative importance of latency and throughput requirements. The environment inherently enforces resource constraints; the reward function primarily focuses on penalizing \ac{QoS} deviations.

Together, these MDP components allow the \ac{DRL} agent to iteratively improve its resource allocation decisions under the complex dynamics of vehicular networks.

\section{Proposed XRL Algorithm}
\label{sec:proposed_xrl}

\begin{figure*}[ht]
\centering
\includegraphics[clip, trim=4cm 6cm 4cm 1.1cm, width=0.8\textwidth]{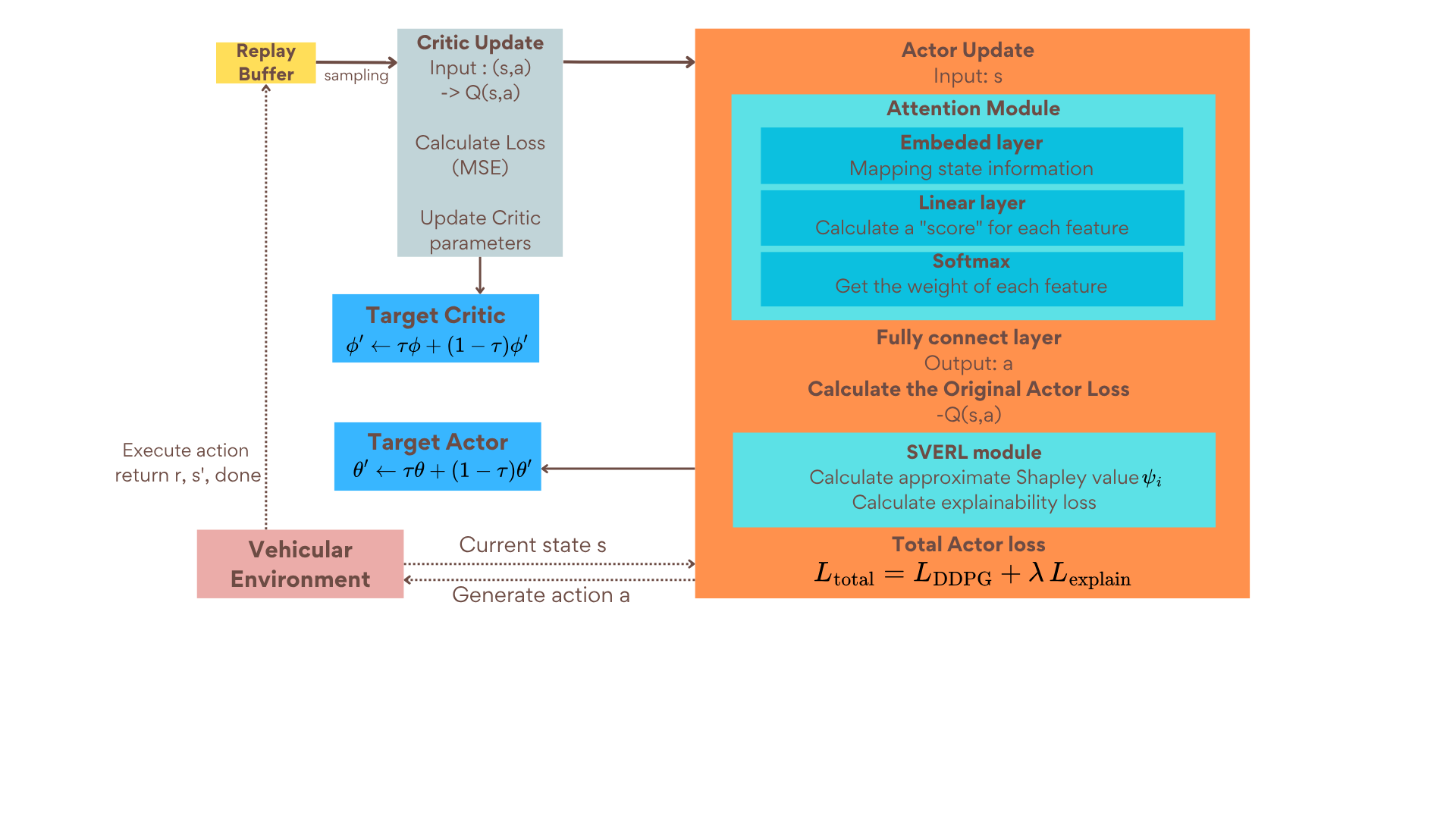}
\caption{Proposed XRL Framework Overview}
\vspace{-0.4cm}

\label{fig:frame}
\end{figure*}

To enhance the transparency and reliability of resource allocation decisions in \ac{V2X} slicing environments, we propose an Explainable Deep Reinforcement learning (XRL) framework that integrates an attention mechanism with offline Shapley value estimation. By integrating our XRL solution into the 
near-RT RIC framework, we aim to achieve both improved system performance and transparent
decision-making, crucial for safety-critical \ac{V2X} scenarios. The Shapley Value for Explaining Reinforcement Learning (SVERL) method introduced in~\cite{beechey2023explaining} rigorously quantifies the contributions of individual state features based on game-theoretic principles; however, due to the need for exhaustive enumeration or Monte Carlo sampling over all feature subsets, the computational overhead of the SVERL method is not suitable for real-time applications such as \ac{V2X}. In contrast, our proposed method employs a lightweight attention module to enable real-time interpretability while leveraging offline Shapley value estimation to ensure theoretical soundness of internal feature importance. This design significantly reduces the computational burden while preserving high explanation fidelity. The overall architecture of the proposed framework is illustrated in Fig.~\ref{fig:frame}.

\subsection{\ac{XAI} Integration}
For a given input state vector $\mathbf{s}\in\mathbb{R}^d$ with 
$d$ representing the total number of state features, our model first computes an attention weight vector via a linear transformation and softmax activation:
\begin{equation}
\boldsymbol{\alpha} = \mathrm{softmax}(W\mathbf{s} + b),
\label{eq:attn}
\end{equation}
where $W\in\mathbb{R}^{d\times d}$ and $b\in\mathbb{R}^{d}$ are trainable parameters. The state vector is then reweighted element-wise as
\begin{equation}
\tilde{\mathbf{s}} = \boldsymbol{\alpha} \odot \mathbf{s},
\label{eq:weighted_state}
\end{equation}
with $\odot$ denoting the Hadamard product. During inference, $\boldsymbol{\alpha}$ directly explains feature importance.

To further enhance interpretability, we approximate the Shapley value for each feature. Given the feature set $F=\{1,2,\dots,d\}$ and any subset $C\subseteq F$, we define a characteristic function based on the expected return when only the features in $C$ are observed:
\begin{equation}
v(C)=\mathbb{E}\Biggl[\sum_{t=0}^{\infty}\gamma^t r_{t+1}\,\Bigg|\, s_0=\mathbf{s},\, \text{observing only } C\Biggr].
\label{eq:char_func}
\end{equation}
The Shapley value for the $i$th feature is then given by
\begin{equation}
\psi_i = \sum_{C\subseteq F\setminus\{i\}} \frac{|C|!(|F|-|C|-1)!}{|F|!}\Bigl[v(C\cup\{i\}) - v(C)\Bigr],
\label{eq:shapley}
\end{equation}
and is approximated via Monte Carlo sampling:
\begin{equation}
\psi_i \approx \frac{1}{M}\sum_{j=1}^{M}\Bigl[v(C_j\cup\{i\}) - v(C_j)\Bigr],
\label{eq:approx_shapley}
\end{equation}
where each $C_j$ is randomly sampled from $F\setminus\{i\}$, and $M$ is the number of samples~\cite{beechey2023explaining}.

To fuse the attention mechanism with the Shapley value estimation, we align their normalized values by defining
\begin{equation}\label{eq:normalized_values}
\hat{\alpha}_i = \frac{\alpha_i}{\sum_{k=1}^{d}\alpha_k}, \quad
\hat{\psi}_i = \frac{\psi_i}{\sum_{k=1}^{d}\psi_k}.
\end{equation}

The auxiliary explanation loss is then defined as
\begin{equation}
L_{\text{explain}} = \frac{1}{d}\sum_{i=1}^{d} \left(\hat{\alpha}_i - \hat{\psi}_i\right)^2.
\label{eq:explain_loss}
\end{equation}
Finally, the overall loss function combines the standard DDPG loss with the explanation loss and is calculated as:
\begin{equation}
L_{\text{total}} = L_{\text{DDPG}} + \lambda \, L_{\text{explain}},
\label{eq:total_loss}
\end{equation}
where $\lambda$ is a hyperparameter controlling the trade-off between performance and explainability.

\begin{algorithm}[!b]

\caption{Attention-DDPG-SVERL}
\label{alg:integrated_sverl_attention}
\KwIn{Environment, networks, replay buffer, batch size $K$, Shapley threshold $\tau$, evaluation interval $N_{\text{eval}}$, explanation loss weight $\lambda$}
\KwOut{Resource allocation decisions with interpretable feature weights}

Initialize environment, networks, replay buffer, and SVERL module\;
Embed an attention layer in the Actor network to output both action $a_t$ and attention weights $\alpha_t$\ at time $t$;

\For{each episode $e=1$ to $E$}{
    Reset environment to get initial state $s_0$\;
    \For{each time step $t=1$ to $T$}{
        $(a_t, \alpha_t) \gets \text{Actor}(s_t)$\;
        Execute $a_t$, observe $s_{t+1}$ and reward $r_t$\;
        Store $\{s_t, a_t, r_t, s_{t+1}\}$ in replay buffer\;
        $s_t \gets s_{t+1}$\;
        
        \If{replay buffer size $\geq K$}{
            Sample mini-batch from buffer and update Critic and Actor\;
            Soft-update target networks\;
            
            \If{$e \mod N_{\text{eval}} = 0$}{
                Sample a subset of states $\{s\}$\;
                For each state, compute approximate Shapley values $\psi$ via Monte Carlo rollouts\;
                Compute explanation loss: $L_{\text{explain}} = \text{MSE}(\text{normalize}(\alpha), \text{normalize}(\psi))$\;
                Update Actor network with total loss: $L_{\text{total}} = L_{\text{DDPG}} + \lambda L_{\text{explain}}$\;
            }
        }
        Add decaying Gaussian noise to $a_t$ and clip to $[0,1]$\;
    }
}

\end{algorithm}

\subsection{O-RAN Near-Real-Time RIC for Resource Management and Control}
Based on the \ac{O-RAN} architecture~\cite{liang2024energy,ahmadi2025towards}, we deploy a near-real-time (near-RT) \ac{RIC} to manage dynamic resource allocation and network slicing for vehicular communications. Unlike conventional scheduling processes that operate at millisecond-level subframe intervals, the near-RT \ac{RIC} performs strategic resource management at coarser intervals, typically on the order of one second~\cite{oran2020usecases}. Specifically, at each decision interval \( t \), the \ac{RIC} observes the global network state \( S_t \) (as defined in ~\eqref{eq:state_definition}) and produces an action \( A_t \) (as described in ~\eqref{eq:action_definition}), which specifies the vehicle-to-\ac{gNB} association variable \( q_{i,m}^t \) and the resource block allocation decision \( B_{i,m}^t \).

These high-level decisions are executed by lower-layer scheduling mechanisms within each decision interval. At the start of every new interval, the near-RT \ac{RIC} collects updated state information reflecting changes in vehicle mobility, channel conditions, and service demand. This continuous feedback loop enables the \ac{RIC} to dynamically re-optimize the slicing and resource allocation strategies.

By aggregating fine-grained millisecond-level scheduling into coarser-interval strategic decisions, this approach significantly reduces the computational complexity of the near-RT \ac{RIC}, maintaining scalability and responsiveness even in rapidly changing network conditions. Moreover, it aligns with the \ac{O-RAN} objective of decoupling control-plane and user-plane functionalities, thereby enabling centralized and efficient network slicing orchestration. 

\section{Results}
We consider a V2X network with an area of 1000m×1000m consisting of 3 \acp{gNB}, and the rest of the settings are shown in Table ~\ref{tab:para}.
\begin{table}[t!]
\centering
\vspace{-0cm}
\caption{List of parameters and values}
\newcolumntype{L}[1]{>{\raggedright\let\newline\\\arraybackslash\hspace{0pt}}p{#1}}
\begin{tabular}{|l|l|}
\hline
\textbf{Parameters} & \textbf{Value/Setting} \\
\hline
 Number of Vehicles & 5 \\
Resource Blocks per BS & 273 \\
Transmission Power & 50dBm \\
Noise Power & $1.4\times10^{-15}$ W \\
Resource Block Bandwidth & 360 kHz \\
Maximum Delay for URLLC & 0.015 sec \\
Minimum Data Rate Requirement for eMBB & 20 Mbps \\
Vehicle Velocity & 15m/s \\
Path Loss Exponent & 3.5 \\
URLLC Data Requirement & $3\times10^{5}$ bits \\
\hline
\end{tabular}
\label{tab:para}
\end{table}  

\subsection{QoS Satisfaction Performance}

\begin{figure}[h!]
\centering
\vspace{-0cm}
\includegraphics[clip, trim=0.0cm 0cm 0.0cm 0cm, width=0.45\textwidth]{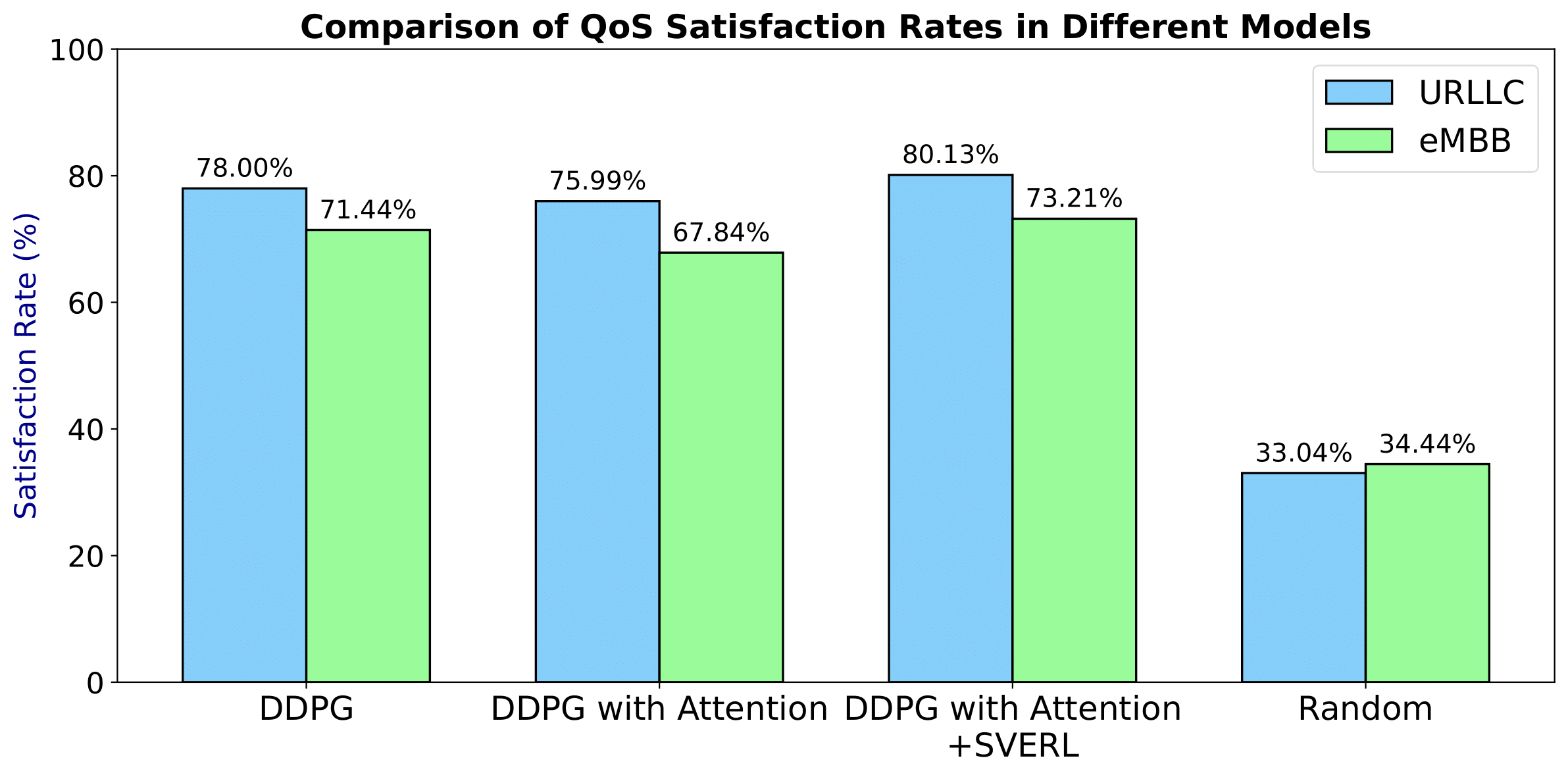}
\caption{QoS Satisfaction Comparison for URLLC and eMBB Services}
\label{fig:qos_satisfaction}
\end{figure}

Fig.~\ref{fig:qos_satisfaction} provides a comparative analysis of QoS satisfaction rates for \ac{URLLC} and \ac{eMBB} services under four different approaches: (i) random algorithm, (ii) standard \ac{DDPG}, (iii) \ac{DDPG} with an attention module, and (iv) our proposed method combining \ac{DDPG} with attention and SVERL. 

In addition to the three learning-based methods, the random baseline achieves only 33.04\% for \ac{URLLC} and 34.44\% for \ac{eMBB}, which is substantially lower than any \ac{RL}-based approach. The results indicate that our proposed combined attention and SVERL mechanism attains the highest \ac{QoS} satisfaction, reaching 80.13\% for \ac{URLLC} and 73.21\% for \ac{eMBB}, marking an effective improvement over the pure \ac{DDPG} model. Furthermore, integrating SVERL with attention outperforms the attention-augmented \ac{DDPG} variant, implying that the explainability component provides transparency and bolsters overall performance.


\subsection{Explainability Analysis}
\begin{figure}[h!]
\centering
\includegraphics[width=0.5\textwidth]{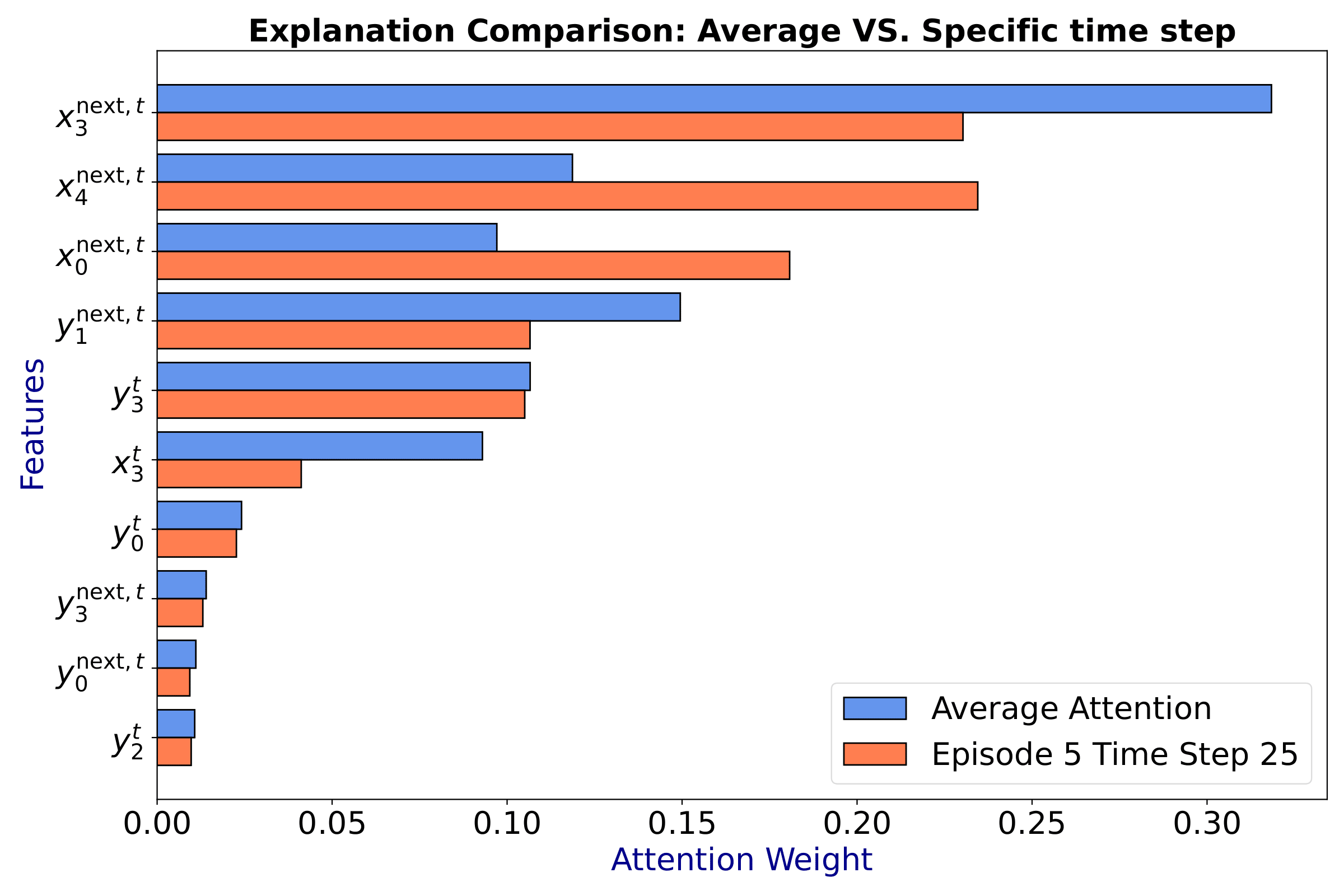}
\caption{Feature Weights Distribution across Test Episodes VS Specific Time Slot}
\label{fig:merged}
\end{figure}

Figure~\ref{fig:merged} presents a merged visualization of the attention weights generated by our attention-\ac{DDPG}-SVERL method. Given the large number of state features and the very low weights for most, we focus only on the ten most significant features for brevity and clarity. The blue panel of the figure shows the average attention weights over the entire test set, while the orange panel displays the attention distribution at a specific time step. 

Here, the features align with the state (as shown in ~\ref{eq:state_definition}). These distributions exhibit a consistent trend in which location-related and predicted mobility features (such as current and next position coordinates) receive higher weights, indicating that the model prioritizes spatial information and mobility patterns to adapt resource allocation decisions in real time. We can directly interpret the agent's behavior by examining the resulting weight values. For instance, a high weight assigned to the next time slot's coordinates reflects the agent’s attempt to adjust proactively to maintain channel condition and minimize \ac{QoS} violations. This emphasis on current and anticipated vehicular positions aligns well with the overarching goal of minimizing \ac{QoS} violations in a dynamic vehicular environment.

\subsection{Explanation Fidelity Evaluation}

To evaluate our proposed model's interpretability and explanation fidelity, we employ the Pearson correlation coefficient as a quantitative metric, measuring the relationship between feature perturbations and the resulting changes in the model’s output \cite{sedgwick2012pearson}. Specifically, we systematically perturb each feature, measure the change in the model’s output, and compare it to the assigned feature importance to assess how well the explanations align with the model's actual behavior. A higher correlation coefficient indicates that the explanations accurately reflect the model’s internal decision-making process.

Table.~\ref{tab:explanation_fidelity} compares the explanation fidelity between a purely attention-based model and our combined attention-SVERL approach. The Pearson correlation coefficient of our integrated attention and SVERL model significantly exceeds the pure attention model. This notable improvement demonstrates that aligning attention weights with Shapley value estimations effectively enhances the transparency and accuracy of feature importance explanations.

The increased fidelity offered by our explainability-enhanced model enables a more reliable and intuitive understanding of the resource management policy --- an essential requirement for trustworthy deployment in safety-critical vehicular communication networks.

\begin{table}[ht]
    \centering
    \caption{Comparison of Explanation Fidelity between Pure Attention and Attention-DDPG-SVERL Models}
    \label{tab:explanation_fidelity}
    \begin{tabular}{lcc}
        \toprule
        \textbf{Method} & \textbf{Mean Correlation} \\
        \midrule
        Attention+SVERL & 0.3054  \\ 
        PureAttention   & 0.1708 \\
        \bottomrule
    \end{tabular}
\end{table}

\section{Conclusion}
In this paper, we proposed an interpretable \ac{DRL} framework for dynamic network slicing and resource allocation in vehicular networks. By integrating an attention mechanism with offline Shapley value estimation, our approach achieves high \ac{QoS} satisfaction for \ac{URLLC} and \ac{eMBB} services, while also enabling real-time interpretability of the decision-making process. Simulation results demonstrate that the proposed method effectively balances resource allocation and explains the model’s behavior transparently. Leveraging theoretically grounded Shapley values, along with an auxiliary loss to align feature importance, further enhances the robustness of the model. Overall, our integrated framework addresses key challenges in dynamic vehicular network management and offers a solid foundation for future research on scalable, interpretable, and reliable resource management solutions. Future work will focus on extending the framework to more complex large-scale and highly mobile urban scenarios to improve real-world deployment capability. Additionally, we plan to leverage the interpretability insights for feature selection to further reduce model complexity and computation overhead.

\begin{acronym} 
\acro{5G}{Fifth Generation}
\acro{6G}{Sixth Generation}
\acro{AI}{Artificial Intelligent }
\acro{ACO}{Ant Colony Optimization}
\acro{ANN}{Artificial Neural Network}
\acro{BB}{Base Band}
\acro{BBU}{Base Band Unit}
\acro{BER}{Bit Error Rate}
\acro{BS}{Base Station}
\acro{BW}{bandwidth}
\acro{C-RAN}{Cloud Radio Access Networks}
\acro{CU}{Central Unit}
\acro{RU}{Radio Unit}
\acro{DU}{Distributed Unit}
\acro{CAPEX}{Capital Expenditure}
\acro{CoMP}{Coordinated Multipoint}
\acro{CR}{Cognitive Radio}
\acro{CRLB}{Cramer-Rao Lower Bound}
\acro{C-RAN}{Cloud Radio Access Network}
\acro{D2D}{Device-to-Device}
\acro{DAC}{Digital-to-Analog Converter}
\acro{DAS}{Distributed Antenna Systems}
\acro{DBA}{Dynamic Bandwidth Allocation}
\acro{DC}{Duty Cycle}
\acro{DFRC}{Dual Function Radar Communication}
\acro{DL}{Deep Learning}
\acro{DSA}{Dynamic Spectrum Access}
\acro{DQL}{Deep Q Learning}
\acro{DQN}{Deep Q-Network}
\acro{DDQN}{double deep Q network}
\acro{DDPG}{Deep Deterministic Policy Gradient}
\acro{FBMC}{Filterbank Multicarrier}
\acro{FEC}{Forward Error Correction}
\acro{FFR}{Fractional Frequency Reuse}
\acro{FL}{Federated Learning}
\acro{FSO}{Free Space Optics}
\acro{NET}{Flying ad-hoc network}
\acro{GA}{Genetic Algorithms}
\acro{GAN}{Generative Adversarial Networks}
\acro{GMMs}{Gaussian mixture models}
\acro{HAP}{High Altitude Platform}
\acro{HL}{Higher Layer}
\acro{HARQ}{Hybrid-Automatic Repeat Request}
\acro{HCA}{Hierarchical Cluster Analysis}
\acro{HO}{Handover}
\acro{KNN}{k-nearest neighbors} 
\acro{IoT}{Internet of Things}
\acro{ISAC}{Integrated Sensing and Communication}
\acro{LAN}{Local Area Network}
\acro{LAP}{Low Altitude Platform}
\acro{LL}{Lower Layer}
\acro{LoS}{Line of Sight}
\acro{LTE}{Long Term Evolution}
\acro{LTE-A}{Long Term Evolution Advanced}
\acro{MAC}{Medium Access Control}
\acro{MAP}{Medium Altitude Platform}
\acro{MDP}{Markov Decision Process}
\acro{ML}{Machine Learning}
\acro{MME}{Mobility Management Entity}
\acro{mmWave}{millimeter Wave}
\acro{MIMO}{Multiple Input Multiple Output}
\acro{NFP}{Network Flying Platform}
\acro{NFPs}{Network Flying Platforms}
\acro{NLoS}{Non-Line of Sight}
\acro{RU}{Radio Unit}
\acro{OFDM}{Orthogonal Frequency Division Multiplexing}
\acro{OSA}{Opportunistic Spectrum Access}
\acro{O-RAN}{Open Radio Access Network}
\acro{C-RAN}{cloud radio access network}
\acro{OMC}{O-RAN Management and Control}
\acro{PAM}{Pulse Amplitude Modulation}
\acro{PAPR}{Peak-to-Average Power Ratio}
\acro{PGW}{Packet Gateway}
\acro{PHY}{physical layer}
\acro{PSO}{Particle Swarm Optimization}
\acro{PU}{Primary User}
\acro{QAM}{Quadrature Amplitude Modulation}
\acro{QoE}{Quality of Experience}
\acro{QoS}{Quality of Service}
\acro{QPSK}{Quadrature Phase Shift Keying}
\acro{RF}{Radio Frequency}
\acro{RIS}{Reconfigurable Intelligent Surce}
\acro{RL}{Reinforcement Learning}
\acro{DRL}{Deep Reinforcement Learning}
\acro{RMSE}{Root Mean Squared Error}
\acro{RN}{Remote Node}
\acro{RRH}{Remote Radio Head}
\acro{RRC}{Radio Resource Control}
\acro{RRU}{Remote Radio Unit}
\acro{RSS}{Received Signal Strength}
\acro{SAR}{synthetic-aperture radar}
\acro{SU}{Secondary User}
\acro{SCBS}{Small Cell Base Station}
\acro{SDN}{Software Defined Network}
\acro{SNR}{Signal-to-Noise Ratio}
\acro{SON}{Self-organising Network}
\acro{SVM}{Support Vector Machine}
\acro{TDD}{Time Division Duplex}
\acro{TD-LTE}{Time Division LTE}
\acro{TDM}{Time Division Multiplexing}
\acro{TDMA}{Time Division Multiple Access}
\acro{TL}{Transfer Learning}
\acro{UE}{User Equipment}
\acro{ULA}{Uniform Linear Array}
\acro{UAV}{Unmanned Aerial Vehicle}
\acro{USRP}{Universal Software Radio Platform}
\acro{XAI}{Explainable AI}
\acro{THz}{Terahertz }
\acro{IRSs}{Intelligent Reflecting Surfaces}
\acro{OAM}{Orbital Angular Momentum}
\acro{LSTM}{Long Short-Term Memory}
\acro{GRU}{Gated Recurrent Unit}
\acro{DNN}{Deep Neural Network}
\acro{RMSE}{Root Mean Square Error}
\acro{SHAP}{SHapley Additive exPlanations}
\acro{SINR}{Signal-to-Interference-plus-Noise Ratio}
\acro{AV}{Atonomous Vehicles}
\acro{CNN}{Convolutional Neural Network}
\acro{VNFs}{Virtual Network Functions}
\acro{DDQL}{Double Deep Q-Learning}
\acro{UDN}{Ultra-Dense Network}
\acro{LIME}{Local Interpretable Model-agnostic Explanations}
\acro{FL}{Federated learning}
\acro{MEC}{ Mobile Edge Computing}
\acro{ACDQL}{Actor-Critic Deep  Q-Learning}
\acro{QL}{ Q-Learning}
\acro{DHOA}{  Deer Hunting Optimisation Algorithm}
\acro{OWFE}{ optimal weighted feature extraction}
\acro{eMBB}{Enhanced Mobile Broadband}
\acro{mMTC}{Massive Machine Type Communications}
\acro{URLLC}{Ultra-Reliable and Low-Latency Communications}
\acro{RRM}{ Radio Resource Management}
\acro{AD}{Anomaly Detection}
\acro{XNN}{EXplainable Neural Network}
\acro{B5G}{Beyond 5G}
\acro{FLaaS}{Federated Learning As A Service}
\acro{IAB}{Integrated Access and Backhaul}
\acro{D3QN}{duel dual DQN}
\acro{MCTS}{Monte Carlo Tree Search}
\acro{XMLAD}{EXplainable Machine Learning for Anomaly Detection and Classification}
\acro{BESS}{Battery Energy Storage Systems}
\acro{IDS}{Intrusion Detection System}
\acro{XRL}{EXplainable Reinforcement Learning}
\acro{LRP}{Layer-by-layer Correlation Propagation}
\acro{DDDQN}{Dual Duel Deep Q-learning neural network}
\acro{xURLLC}{Extreme Ultra-Reliable Low-Latency Communication}
\acro{PER}{Packet Error Rate}
\acro{IG}{Integrated Gradient}
\acro{STRR}{Short-Term Resource Reservation}
\acro{KPI}{Key Performance Indicator}
\acro{SOC}{State-Of-Charge}
\acro{CLC}{closed-loop control}
\acro{RIC}{RAN Intelligent controller}
\acro{CPU}{Central Processing Unit}
\acro{SLOT}{Self-Learning Optimization Training}
\acro{NAT}{Network Address Translation}
\acro{IoE}{Internet of Everything}
\acro{SLA}{Service Level Agreement}
\acro{G5IAD}{Graph-based Interpretable Anomaly Detection}
\acro{FDL}{Federated deep learning}
\acro{PIRL}{Pseudo-Inverse Relevance Learning}
\acro{V2X}{Vehicle-to-Everything}
\acro{QoT}{Quality of Trust}
\acro{IIoT}{Industrial Internet of Thing}
\acro{AMC}{Automatic Modulation Classification}
\acro{SMC}{Sparse Multinomial Classifier}
\acro{4G}{4th Generation}
\acro{PI}{Permutation Feature Importance}
\acro{FNN}{Feedforward Neural Network}
\acro{SAGIN}{Space-Air-Ground Integrated Network}
\acro{RAN}{Radio Access Network}
\acro{V2V}{Vehicle-to-Vehicle}
\acro{V2I}{Vehicle-to-Infrastructure}
\acro{V2P}{Vehicle-to-Pedestrian}
\acro{ITS}{Intelligent Transportation Systems}
\acro{DSRC}{Dedicated Short-Range Communication}
\acro{MLOps}{Machine Learning Operations}
\acro{PDP}{Partial Dependence Plot}
\acro{RuleFit}{Rule-based Model Fitting}
\acro{NOMA}{Non-Orthogonal Multiple Access}
\acro{gNB}{gNodeB}
\acro{PRB}{Physical Resource Block}
\end{acronym}

\bibliographystyle{IEEEtran}
\bibliography{References.bib}
\end{document}